# Meteorological time series forecasting with pruned multi-layer perceptron and 2-stage Levenberg-Marquardt method

cyril voyant[1*], Wani Tamas[1], Marie-Laure Nivet[1], Gilles Notton[1], Christophe Paoli[2], Aurélia Balu[1], Marc Muselli[1]

1-University of Corsica, UMR CNRS 6134 SPE, Campus Grimaldi, BP 52, 20250 Corte, (France)

2- Departement de Genie Informatique, Universite Galatasaray. No:36 34357 Ortakoy, Istanbul, Turkey

*corresponding author; tel +33495293666, fax +33495293797, cyril.voyant@ch-castelluccio.fr

**Abstract:** A Multi-Layer Perceptron (MLP) defines a family of artificial neural networks often used in TS modeling and forecasting. Because of its "black box" aspect, many researchers refuse to use it. Moreover, the optimization (often based on the exhaustive approach where "all" configurations are tested) and learning phases of this artificial intelligence tool (often based on the Levenberg-Marquardt algorithm; LMA) are weaknesses of this approach (exhaustively and local minima). These two tasks must be repeated depending on the knowledge of each new problem studied, making the process, long, laborious and not systematically robust. In this paper a pruning process is proposed. This method allows, during the training phase, to carry out an inputs selecting method activating (or not) inter-nodes connections in order to verify if forecasting is improved. We propose to use iteratively the popular damped least-squares method to activate inputs and neurons. A first pass is applied to 10% of the learning sample to determine weights significantly different from 0 and delete other. Then a classical batch process based on LMA is used with the new MLP. The validation is done using 25 measured meteorological TS and cross-comparing the prediction results of the classical LMA and the 2-stage LMA.

## 1. Background

The primary goal of time series (TS) analysis is forecasting, i.e. using the past to predict the future (Bourbonnais, 1998; Faraday and Chatfield, 1998; Georgakarakos et al., 2006; Hamilton, 1994; Voyant et al., 2013, 2009; Wang et al., 2011, 2013; Xie et al., 2006). This formalism is used in many scientific fields like econometrics, seismology or meteorology. Lot of methods are dedicated to the prediction of discrete phenomena, one of the most popular is the artificial neural network (ANN) (Ali, 2013; De Gooijer and Hyndman, 2006; Voyant et al., 2012b). From a mathematical point of view, ANN is a function defined as the composition of other functions (Cybenko, 1989). Members of the class of such functions are obtained by varying parameters, (as connections or weights) (Lauret et al., 2008). A Multi-Layer Perceptron (MLP) defines a family of functions often used in TS modeling (Voyant et al., 2012). In this model, neurons are grouped in layers and only forward connections exist. A typical MLP consists of an input, hidden and output layers, including neurons, weights and a transfer functions (Crone, 2005; Wang et al., 2010). Each neuron (noted $i$) transforms the weighted sum (weight $w_{ij}$, bias $b_i$) of inputs ($x_j$) into an output ($y_i = f(\sum_{j=1}^{n} x_j w_{ij} + b_i)$) using a transfer function ($f$). The goal of this method is to determine weights and bias for a given problem. A complex process is necessary to adapt connections using a suitable training algorithm (often based on the Levenberg-Marquardt algorithm; LMA (Fan and Pan, 2009; Yan et al., 2014; Yoo et al., 2003). The training step is dependent of the number of inputs, layers and hidden nodes. The better configuration defines the optimized MLP (Voyant et al., 2012). This step is the weaknesses of this approach because no consensus or scientific rules exist, often the use of the exhaustive approach (where "all" configurations are tested) is the only usable and must be repeated for new studied problem, making the process, long, laborious and not systematically robust. This "black box" aspect leads any researchers to refuse to use it. In this paper a pruning process allowing to automatically selecting inputs is proposed. This method allows, during the training phase, to carry out a selecting method activating (or not) inter-nodes connections (More, 2003). With this process, the optimization step becomes self-acting and the parsimony principle is kept. Less the MLP is complex more it is efficient (Cybenko, 1989).

## 2. Materials and methods

We propose to use iteratively the popular LMA also known as the damped least-squares method to activate the inputs and neurons (*m* weights and bias) (Brusset et al., 1976). A first pass is applied to 10% of the learning sample. For each step, a system of *m* non-linear equations with *m* unknowns is solved (see equation 1 in case of 1 hidden layer MLP where *O* and *I* are the outputs and inputs, $W^1$, $B^1$ and $W^2$, $B^2$ the weights and bias matrices of the hidden and output layers) (Crone and Kourentzes, 2010).

$$O^i = W^2 \cdot (\tanh(W^1 \cdot I^i + B^1)) + B^2 \quad \text{with } 1 < I < m \quad (1)$$

After this first phase, each weights and bias ($\omega_{i \in [1,m]}$) are represented by probability distributions. A statistical test based on the bootstrap distribution is used to determine if the first moment of each $\omega_i$ is significantly different from zero (Kreiss and Paparoditis, 2011). Before to initiate the second pass, the network is customized and connections related to each $\omega_i$ non-significantly different from 0 are canceled. Then a classical batch process based on LMA is used with the new MLP. We validated our method with five hourly meteorological TS (wind direction *WD*, wind speed *WS*, Global radiation *Glo*, Humidity *Hum* and temperature *Tem*), each one measured in 5 French sites (Ajaccio, Bastia, Corte, Marseille and Nice) (Troccoli, 2010; Voyant et al., 2012). Note that no pretreatment are operated and that the different TS are not necessary made stationary (periodic TS). For all TS and locations, we used 3200 measures for the training and 400 for the cross comparison between the classical LMA and the 2-stage LMA during the year 2008.

### 2.1. First pass

The first stage begins with the generation of *N* (10% of the total data used during training randomly chosen) systems of *m* nonlinear equations with *m* unknowns (MLP constructed with *m* weights and bias). The method chosen for solve this problem is the LMA method and the ad-hoc objective function *F* (mean square error between calculations and measures). It is an approximation of the Gauss-newton method, the result of the $k^{th}$ iterations ($\omega_k$) corresponding to the local minimum of the function *F* is generated by the linear set of equations (1) (Brusset et al., 1976; Dias et al., 2006; Fan and Pan, 2009; Yoo et al., 2003):

$$(J(\omega_k)^T \cdot J(\omega_k) + \lambda_k I) \cdot d_k = -J(\omega_k)^T \cdot F(\omega_k) \quad (2)$$

*J* denotes the Jacobian matrix of *F* and the scalar $\lambda_k$ controls both the magnitude and variation $d_k$ ($d_k = \Delta\omega = \omega_{k+1} - \omega_k$). After the *N* solving, all the weights and bias are represented by a distribution which will be used during the second stage of the second stage of the methodology. Note that these distributions are not normal (according to the Jarque-Bera test).

### 2.2. Second pass

Before to initiate the second pass, the network is customized and connections related to each $\omega_i$ non-significantly different from zero are canceled. An example of the distribution related to one weight is available in the Figure 1

**Figure 1: example of a MLP weight distribution**

To perform the second pass, we use confidence interval from the bootstrap distributions (4000 samples) of the weights and bias parameters (Kreiss and Paparoditis, 2011). The rule of selection (directly linked to the $\alpha$ value of significance model) is based on the product of the two endpoints $t_1$ and $t_2$ defined respectively by the $\alpha/2^{th}$ and $(1-\alpha/2)^{th}$ percentiles of the distribution. If $t_1.t_2 < 0$ the weight (or bias) is considered as non-significantly different from zero, else it is considered different from zero. The connection of the MLP corresponding to the first case ($t_1.t_2 < 0$) are cancelled, others are kept. The pruned MLP (noted pMLP) is then trained with the classical LMA.

## 3. Results

In the Figure 2a is represented the box plot of the nRMSE (i.e. normalized root mean square error (Voyant et al., 2014)) distribution concerning the five meteorological parameters and the five studied cities. For each case, seven runs are operated, so 175 manipulations are performed with the pruned methodology described above (at left in the Figure 2 and noted pMLP) and the standard approach (at right in the Figure 2 and noted MLP). The chosen architecture is the same for all cases: 7 inputs representing the seven first lags of the meteorological parameter tested and 2 hidden nodes (only one hidden layer). In this figure, we see that the first, the second and the third quartiles are equivalent; thereby we understand that a lot of connections and weights are superfluous. The results of all the simulations are represented in the Figure 1b. Only the points positioned in the upper zone are related to "pMLP is better than MLP" cases are plotted. It appears the points are closer to the y=1 curve in the top area rather than in the bottom area, but this observation seems insignificant.

**Figure 2: a. nRMSE distribution comparison related to pMLP (at left) and MLP (at right), b. ratio of the nRMSE generated by MLP and pMLP**

The table 1 and 2 expose the results for all locations and parameters of the MLP and pMLP approach, the minimum of the nRMSE and nMAE (normalizes mean square error) (Voyant et al., 2012) through the seven runs are exposed in the first table and the averages in the second.

**Table 1: nRMSE and nMAE minima for all location and all parameters, in bold the better results between MLP and pMLP**

**Table 2: nRMSE and nMAE mean for all location and all parameters, in bold the better results between MLP and pMLP**

If the mean of the error metrics is interesting to compare the global trend, the minimum values allow to determine the best learning, and therefore the best network. pMLP is very slightly better than MLP, in the minima case, in 60% of the cases, the nMRSE and the nMAE related to the pMLP are the lowest. Note that the pruning concerns about 20% of weights and bias. In the Figure 3, is available the predictions related to the five time series compares to the measurements.

**Figure 3: Profile during 100 days concerning measurements (lines) MLP (cross) and pMLP (circles)**

In this figure, pMLP gives better visual results for temperature, however concerning the four other time series the gain is not significant.

## 4. Conclusions

The 2-pass approach improves slightly the forecasting quality. In average, 20% of the connections are removed with this approach. According to the parsimony principle, these simplifications increase the generalization capacity and should allow building a robust predictor (Mellit et al., 2009).

This first study done in a quasi-optimized case (7 inputs and 2 hidden neurons) precedes a more general one, where a standard MLP (more than 15 inputs and 15 hidden nodes) will be studied. Indeed, we have shown that the pruning method presented here is able to simplify the network while the performance is roughly equivalent. Applying the 2-pass approach in a 15x15 MLP should allow to optimize it without apply the exhaustive test where all the architectures are try out. Moreover, for users it is a totally transparent methodology, suitable for all TS and faster than classical optimization process. According to the conclusion of this study, it possible that results based on the 2-stage approach may be better than the classical approach based on the "1-stage" LM algorithm.

## 5. References

Ali, M.M.I., 2013. Efficiency optimisation with PI gain adaptation of field-oriented control applied on five phase induction motor using AI technique. Int. J. Model. Identif. Control 20, 344. doi:10.1504/IJMIC.2013.057568

Bourbonnais, R., 1998. Analyse des séries temporelles en économie. Presses Universitaires de France - PUF.

Brusset, H., Depeyre, D., Petit, J.-P., Haffner, F., 1976. On the convergence of standard and damped least squares methods. J. Comput. Phys. 22, 534–542. doi:10.1016/0021-9991(76)90048-6

Crone, S.F., 2005. Stepwise Selection of Artificial Neural Networks Models for Time Series Prediction. Journal of Intelligent Systems 15.

Crone, S.F., Kourentzes, N., 2010. Feature selection for time series prediction – A combined filter and wrapper approach for neural networks. Neurocomputing 73, 1923–1936. doi:10.1016/j.neucom.2010.01.017

Cybenko, G., 1989. Approximation by superpositions of a sigmoidal function. Math. Control Signals Syst. 2, 303–314. doi:10.1007/BF02551274

Dias, F.M., Antunes, A., Vieira, J., Mota, A., 2006. A sliding window solution for the on-line implementation of the Levenberg–Marquardt algorithm. Eng. Appl. Artif. Intell. 19, 1–7. doi:10.1016/j.engappai.2005.03.005

Fan, J., Pan, J., 2009. A note on the Levenberg–Marquardt parameter. Appl. Math. Comput. 207, 351–359. doi:10.1016/j.amc.2008.10.056

Faraday, J., Chatfield, C., 1998. Times Series Forecasting with Neural Networks: A Case Study. Applied Statistics.

Georgakarakos, S., Koutsoubas, D., Valavanis, V., 2006. Time series analysis and forecasting techniques applied on loliginid and ommastrephid landings in Greek waters. Fish. Res. 78, 55–71. doi:10.1016/j.fishres.2005.12.003

Hamilton, J., 1994. Time series analysis. Princeton University Press, Princeton N.J.

Kreiss, J.-P., Paparoditis, E., 2011. Bootstrap methods for dependent data: A review. J. Korean Stat. Soc. 40, 357–378. doi:10.1016/j.jkss.2011.08.009

Lauret, P., Fock, E., Randrianarivony, R.N., Manicom-Ramsamy, J.F., 2008. Bayesian neural network approach to short time load forecasting. Energy Convers. Manag. 49, 1156–1166.

Mellit, A., Kalogirou, S.A., Hontoria, L., Shaari, S., 2009. Artificial intelligence techniques for sizing photovoltaic systems: A review. Renew. Sustain. Energy Rev. 13, 406–419. doi:10.1016/j.rser.2008.01.006

More, A., 2003. Forecasting wind with neural networks. Mar. Struct. 16, 35–49. doi:10.1016/S0951-8339(02)00053-9

Troccoli, A., 2010. Seasonal climate forecasting. Meteorol. Appl. 17, 251–268. doi:10.1002/met.184

Voyant, C., Muselli, M., Paoli, C., Nivet, M.-L., 2012. Numerical weather prediction (NWP) and hybrid ARMA/ANN model to predict global radiation. Energy 39, 341–355. doi:10.1016/j.energy.2012.01.006

Voyant, C., Muselli, M., Paoli, C., Nivet, M.-L., Poggi, P., Haurant, P., 2009. Predictability of PV power grid performance on insular sites without weather stations : use of artificial neural networks. Presented at the 24th European Photovoltaic Solar Energy Conference and Exhibition, Hambourg.

Voyant, C., Notton, G., Paoli, C., Nivet, M.L., Muselli, M., Dahmani, K., 2014. Numerical weather prediction or stochastic modeling: an objective criterion of choice for the global radiation forecasting. ArXiv14016002 Cs Stat.

Voyant, C., Paoli, C., Muselli, M., Nivet, M.-L., 2013. Multi-horizon solar radiation forecasting for Mediterranean locations using time series models. Renew. Sustain. Energy Rev. 28, 44–52. doi:10.1016/j.rser.2013.07.058

Wang, B.S., Ni, Y.Q., Ko, J.M., 2011. Damage detection utilising the artificial neural network methods to a benchmark structure. Int. J. Struct. Eng. 2, 229. doi:10.1504/IJSTRUCTE.2011.040782

Wang, J., Zhu, S., Zhang, W., Lu, H., 2010. Combined modeling for electric load forecasting with adaptive particle swarm optimization. Energy 35, 1671–1678. doi:doi: DOI: 10.1016/j.energy.2009.12.015

Wang, L., Hu, P., Lu, J., Chen, F., Hua, Q., 2013. Neural network and PSO-based structural approximation analysis for blade of wind turbine. Int. J. Model. Identif. Control 18, 69. doi:10.1504/IJMIC.2013.051936

Xie, J.X., Cheng, C.T., Chau, K.W., Pei, Y.Z., 2006. A hybrid adaptive time-delay neural network model for multi-step-ahead prediction of sunspot activity. Int. J. Environ. Pollut. 28, 364. doi:10.1504/IJEP.2006.011217

Yan, C., Li, Y., Zhang, J., 2014. Adaptive neural network control for a class of strict feedback non-linear time delay systems. Int. J. Model. Identif. Control 21, 401. doi:10.1504/IJMIC.2014.062027

Yoo, C.K., Sung, S.W., Lee, I.-B., 2003. Generalized damped least squares algorithm. Comput. Chem. Eng. 27, 423–431. doi:10.1016/S0098-1354(02)00219-3

## 6. Figures

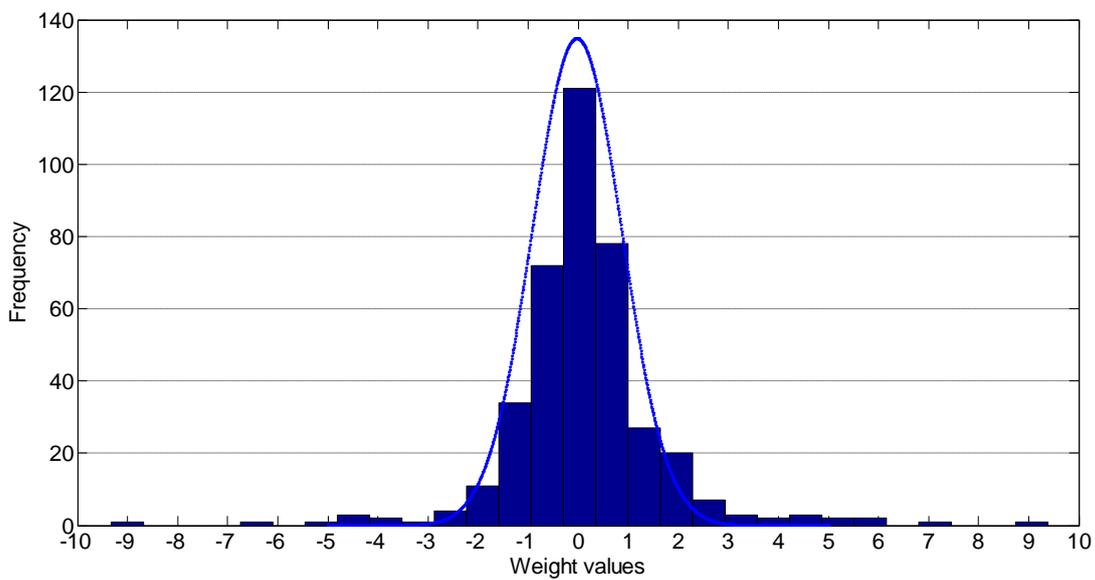

**Figure 1: example of a MLP weight distribution**

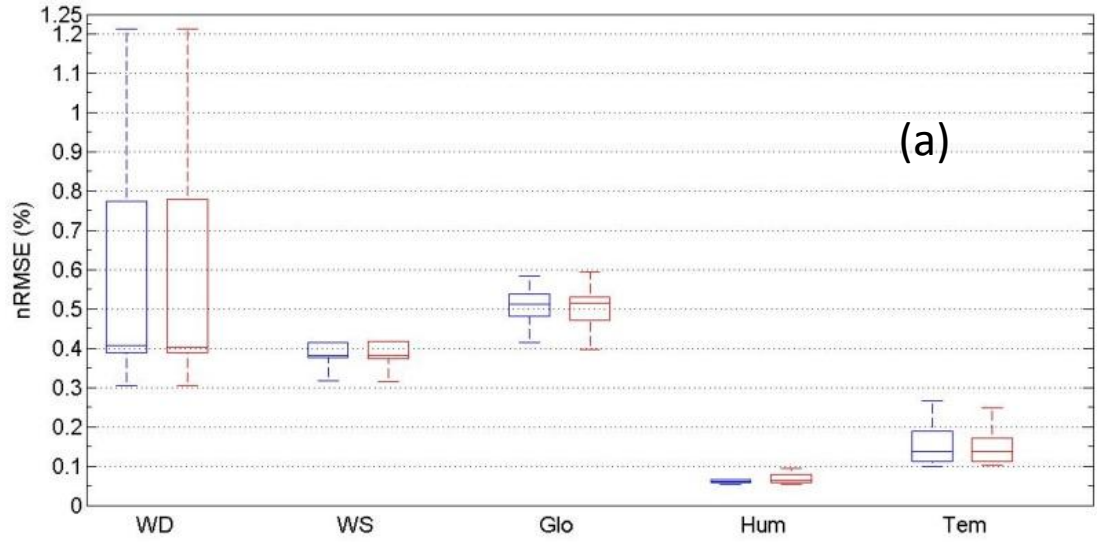

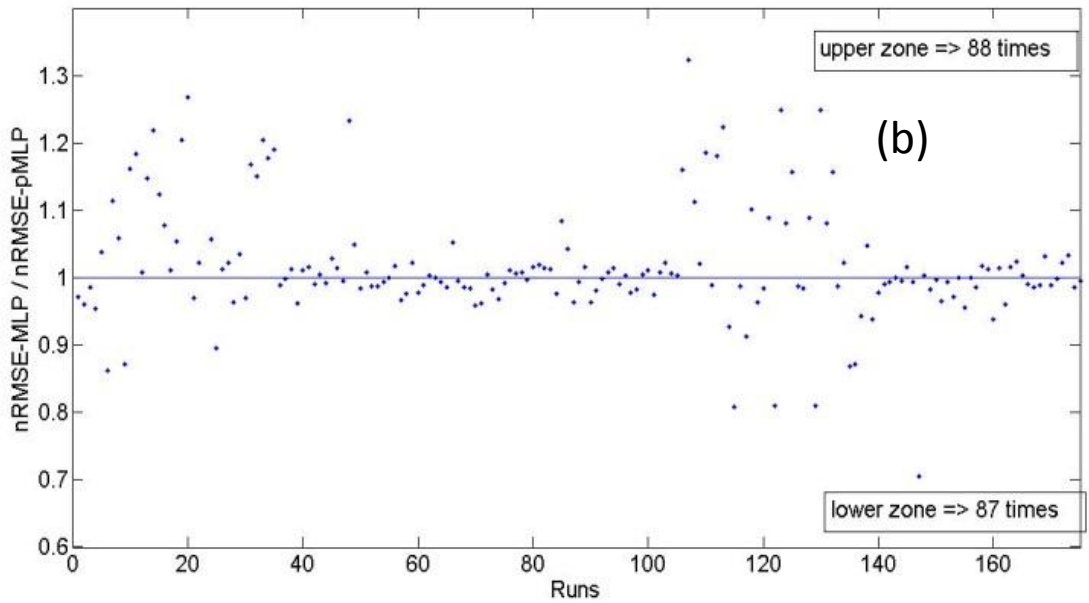

Figure 2: a. nRMSE distribution comparison related to pMLP (at left) and MLP (at right), b. ratio of the nRMSE generated by MLP and pMLP

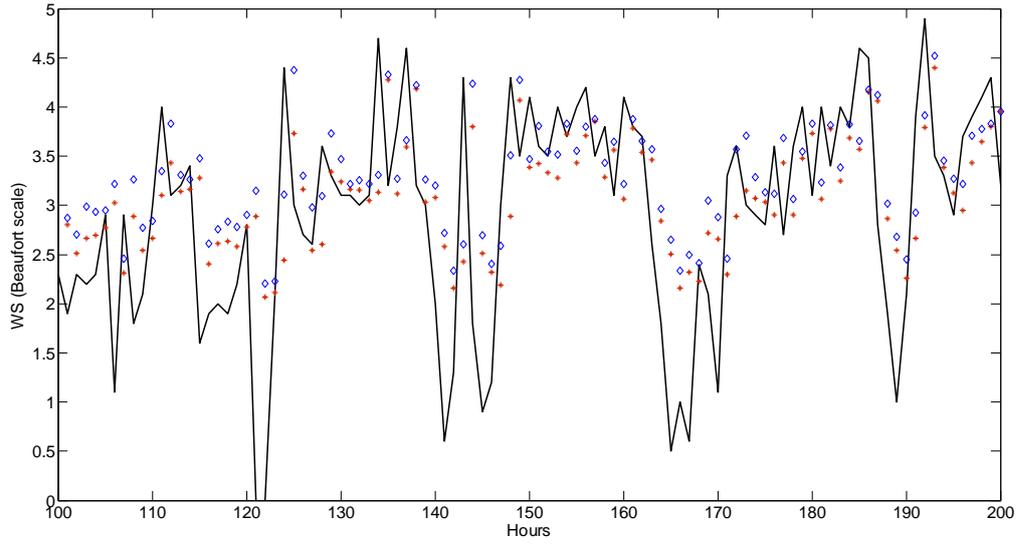

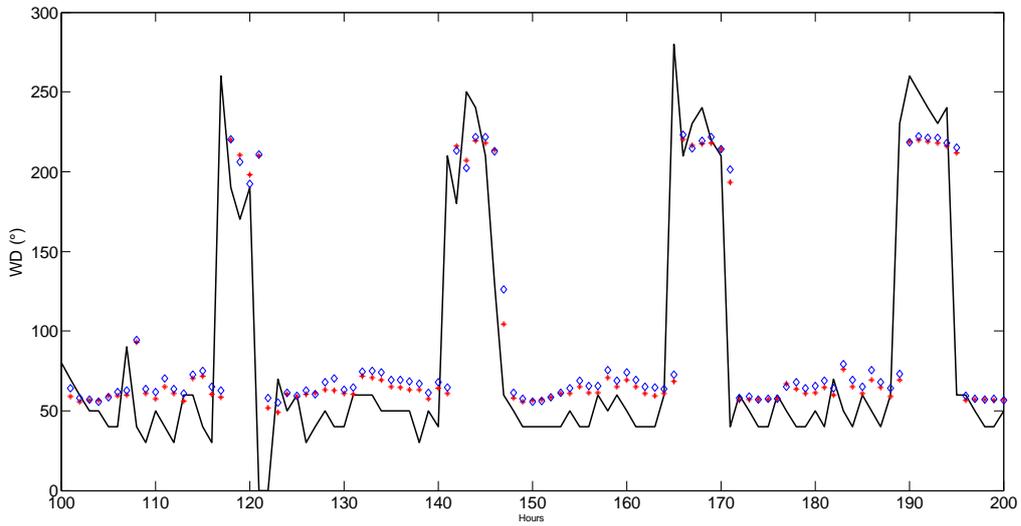

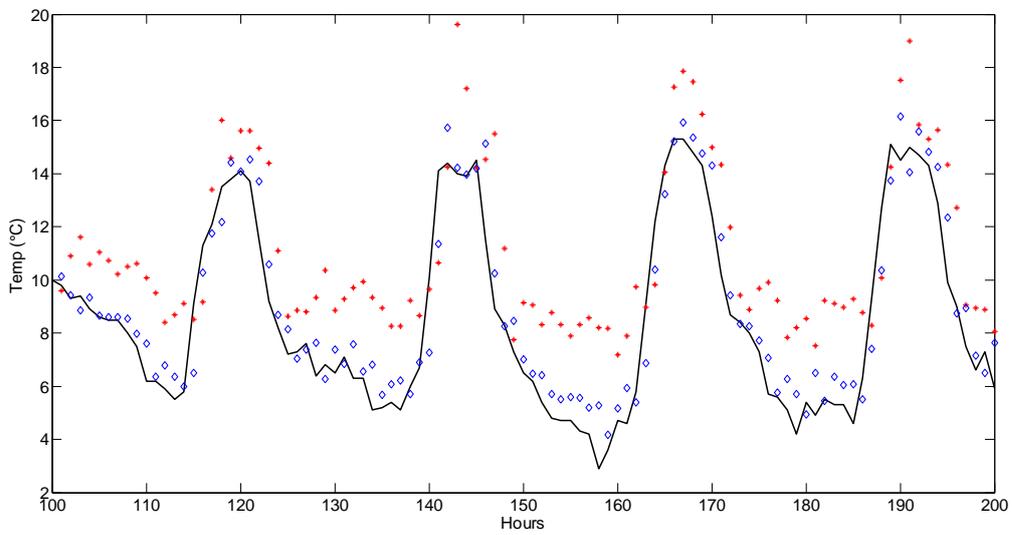

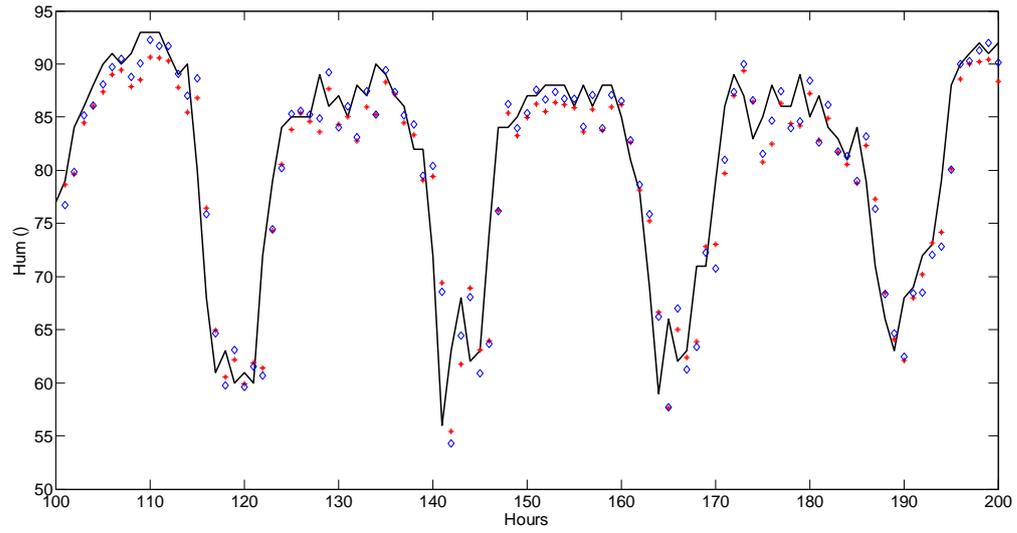

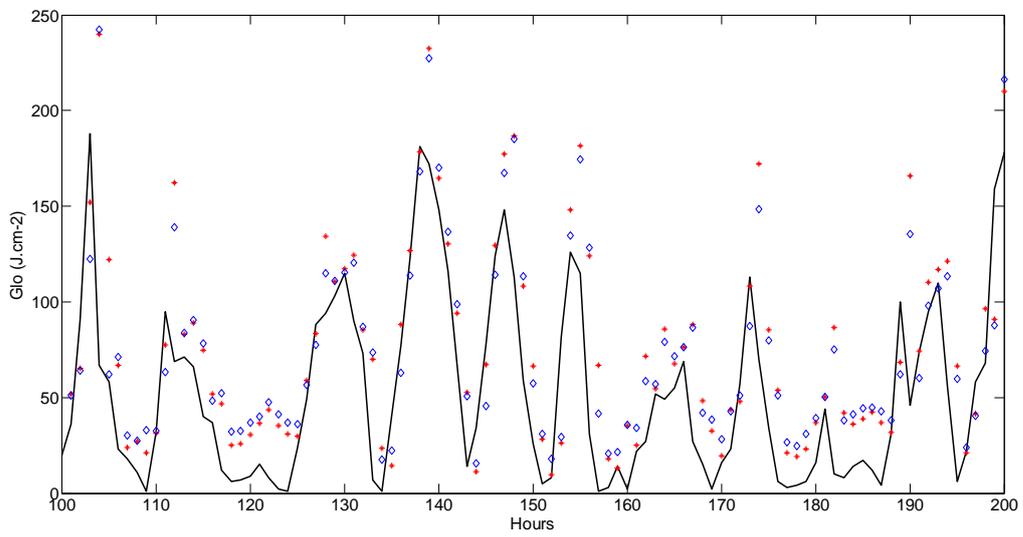

**Figure 3: Profile during 100 days concerning measurements (lines) MLP (cross) and pMLP (circles)**

# 7. Tables

| Data | City | MLP | | pMLP | | |
|---|---|---|---|---|---|---|
| | | *nRMSE* | *nMAE* | Ratio pruning | *nRMSE* | *nMAE* |
| WD | *Aja* | 0.765 | 0.463 | 0.27 | **0.762** | **0.461** |
| | *Bas* | 0.393 | 0.257 | 0.27 | **0.388** | **0.252** |
| | *Cor* | **1.191** | 0.941 | 0.21 | 1.194 | **0.931** |
| | *Mar* | **0.387** | **0.256** | 0.20 | 0.388 | 0.257 |
| | *Nic* | **0.304** | **0.199** | 0.20 | 0.305 | 0.211 |
| WS | *Aja* | **0.399** | **0.302** | 0.11 | 0.410 | 0.308 |
| | *Bas* | 0.374 | **0.277** | 0.18 | **0.370** | 0.279 |
| | *Cor* | **1.148** | 0.910 | 0.22 | 1.183 | **0.902** |
| | *Mar* | **0.377** | 0.264 | 0.21 | **0.377** | **0.263** |
| | *Nic* | 0.318 | 0.206 | 0.23 | **0.314** | **0.202** |
| Glo | *Aja* | 0.525 | 0.413 | 0.17 | **0.472** | **0.372** |
| | *Bas* | **0.439** | **0.323** | 0.21 | 0.456 | 0.370 |
| | *Cor* | **0.298** | **0.214** | 0.21 | 0.323 | 0.250 |
| | *Mar* | 0.416 | 0.346 | 0.21 | **0.378** | **0.302** |
| | *Nic* | **0.455** | **0.380** | 0.20 | 0.465 | 0.386 |
| Hum | *Aja* | **0.064** | 0.049 | 0.22 | **0.064** | **0.048** |
| | *Bas* | **0.061** | **0.044** | 0.20 | **0.061** | 0.045 |
| | *Cor* | **0.055** | **0.036** | 0.23 | 0.056 | 0.037 |
| | *Mar* | **0.053** | **0.036** | 0.23 | **0.053** | **0.036** |
| | *Nic* | **0.083** | **0.059** | 0.19 | **0.083** | **0.059** |
| Tem | *Aja* | **0.099** | **0.077** | 0.29 | 0.101 | **0.077** |
| | *Bas* | 0.113 | 0.080 | 0.25 | **0.111** | **0.079** |
| | *Cor* | 0.224 | 0.158 | 0.29 | **0.207** | **0.133** |
| | *Mar* | 0.111 | 0.079 | 0.28 | **0.109** | **0.073** |
| | *Nic* | 0.147 | **0.108** | 0.21 | **0.146** | 0.111 |

**Table 2: nRMSE and nMAE minima for all location and all parameters, in bold the better results between MLP and pMLP**

| Data | City | MLP | | pMLP | | |
|---|---|---|---|---|---|---|
| | | *nRMSE* | *nMAE* | Ratio pruning | *nRMSE* | *nMAE* |
| WD | *Aja* | **0.772** | **0.478** | 0.27 | 0.774 | 0.484 |
| | *Bas* | 0.418 | 0.287 | 0.27 | **0.401** | **0.270** |
| | *Cor* | **1.197** | 0.972 | 0.21 | 1.202 | **0.958** |
| | *Mar* | **0.391** | **0.262** | 0.20 | 0.394 | 0.269 |
| | *Nic* | **0.308** | **0.212** | 0.20 | 0.310 | 0.218 |
| WS | *Aja* | **0.409** | **0.309** | 0.11 | 0.414 | 0.312 |
| | *Bas* | 0.378 | **0.280** | 0.18 | **0.376** | 0.281 |
| | *Cor* | 1.203 | 0.960 | 0.22 | **1.196** | **0.915** |
| | *Mar* | **0.379** | **0.266** | 0.21 | 0.380 | **0.266** |
| | *Nic* | 0.323 | 0.212 | 0.23 | **0.322** | **0.210** |
| Glo | *Aja* | 0.574 | 0.465 | 0.17 | **0.562** | **0.400** |
| | *Bas* | 0.551 | **0.453** | 0.21 | **0.386** | 0.467 |
| | *Cor* | **0.389** | **0.304** | 0.21 | 0.464 | 0.311 |
| | *Mar* | **0.481** | 0.410 | 0.21 | 0.514 | **0.393** |
| | *Nic* | 0.489 | **0.413** | 0.20 | **0.400** | 0.434 |
| Hum | *Aja* | **0.069** | **0.054** | 0.22 | 0.073 | 0.056 |
| | *Bas* | **0.061** | **0.045** | 0.20 | 0.062 | 0.046 |
| | *Cor* | **0.057** | **0.039** | 0.23 | 0.058 | **0.039** |
| | *Mar* | **0.054** | **0.037** | 0.23 | 0.054 | 0.037 |
| | *Nic* | 0.085 | 0.061 | 0.19 | **0.084** | **0.060** |
| Tem | *Aja* | **0.109** | **0.087** | 0.29 | 0.112 | 0.089 |
| | *Bas* | 0.139 | 0.111 | 0.25 | **0.128** | **0.102** |
| | *Cor* | 0.259 | 0.205 | 0.29 | **0.236** | **0.176** |
| | *Mar* | **0.112** | **0.081** | 0.28 | 0.126 | 0.090 |
| | *Nic* | 0.179 | 0.138 | 0.21 | **0.159** | **0.121** |

**Table 2: nRMSE and nMAE mean for all location and all parameters, in bold the better results between MLP and pMLP**

## 8. Keywords